\documentclass[journal]{IEEEtai}
\usepackage[colorlinks,urlcolor=blue,linkcolor=blue,citecolor=blue]{hyperref}
\usepackage{color,array}
\usepackage{graphicx}
\usepackage{cite}
\usepackage{hyperref}
\usepackage{url}
\usepackage{amsmath}
\usepackage{subcaption}
\let\labelindent\relax
\usepackage{enumitem}
\usepackage{array}
\usepackage{tabularx}
\usepackage[utf8]{inputenc}
\usepackage{amsfonts}
\usepackage{amsmath}
\usepackage[utf8]{inputenc}
\usepackage[T1]{fontenc}
\usepackage{tikz,xcolor,hyperref}
\usepackage{amsthm}
\definecolor{lime}{HTML}{A6CE39}
\DeclareRobustCommand{\orcidicon}{%
	\begin{tikzpicture}
	\draw[lime, fill=lime] (0,0) 
	circle [radius=0.16] 
	node[white] {{\fontfamily{qag}\selectfont \tiny ID}};
	\draw[white, fill=white] (-0.0625,0.095) 
	circle [radius=0.007];
	\end{tikzpicture}
	\hspace{-2mm}
}
\foreach \x in {A, ..., Z}{%
	\expandafter\xdef\csname orcid\x\endcsname{\noexpand\href{https://orcid.org/\csname orcidauthor\x\endcsname}{\noexpand\orcidicon}}
}

\usepackage{lettrine}
\newcommand*{\Resize}[2]{\resizebox{#1}{!}{$#2$}}%

\theoremstyle{definition}
\newtheorem{definition}{Definition}[section]

\begin{document}

\title{{\sc The Fairness Stitch}: Unveiling the Potential of Model Stitching in Neural Network De-Biasing}

\author{{Modar Sulaiman\orcidA{}} and {Kallol Roy\orcidB{}}

\thanks{Modar Sulaiman and Kallol Roy are with the Institute of Computer Science, University of Tartu, 50090 Tartu, Estonia (e-mail: modar.sulaiman@ut.ee; kallol.roy@ut.ee).}
}

\maketitle

\begin{abstract}
The pursuit of fairness in machine learning models has emerged as a critical research challenge in different applications ranging from bank loan approval to face detection. Despite the widespread adoption of artificial intelligence algorithms across various domains, concerns persist regarding the presence of biases and discrimination within these models. To address this pressing issue, this study introduces a novel method called  "The Fairness Stitch (TFS)" to enhance fairness in deep learning models. This method combines model stitching and training jointly, while incorporating fairness constraints. In this research, we assess the effectiveness of our proposed method by conducting a comprehensive evaluation of two well-known datasets, CelebA and UTKFace. We systematically compare the performance of our approach with the existing baseline method. Our findings reveal a notable improvement in achieving a balanced trade-off between fairness and performance, highlighting the promising potential of our method to address bias-related challenges and foster equitable outcomes in machine learning models. This paper poses a challenge to the conventional wisdom of the effectiveness of the last layer in deep learning models for de-biasing. \footnote{Preprint: Code is available at \url{https://github.com/Modar7/The_Fairness_Stitch}}

\end{abstract}

\begin{IEEEImpStatement}
The advent of machine learning has revolutionized diverse sectors, yet concerns persist regarding biases within algorithms. Addressing this, we present "The Fairness Stitch," an innovative framework combining model stitching and training with explicit fairness constraints. Our study rigorously evaluates this method on the CelebA and UTKFace datasets, comparing it with existing baseline method. Results showcase significant improvement in balancing fairness and performance, marking a crucial contribution to fairness-aware computing. "The Fairness Stitch" not only introduces a practical solution but also extends the boundaries of knowledge in fairness-aware machine learning, promising equitable outcomes while preserving model accuracy. This research is pivotal for fostering fairness in machine learning applications, a critical step toward mitigating societal biases.
\end{IEEEImpStatement}

\begin{IEEEkeywords}
Artificial intelligence (AI), AI bias, deep learning, fairness in machine learning, finetune, model stitching, overfitting.
\end{IEEEkeywords}

\section{Introduction}
\IEEEPARstart{T}{he} widespread adoption of machine learning algorithms has transformed numerous sectors, ranging from healthcare and finance to education and criminal justice. Machine learning algorithms have demonstrated exceptional capabilities, enabling data-driven decision-making at an unprecedented scale. However, the increased reliance on automated systems has raised concerns about fairness, as biases and discrimination can be inadvertently embedded in algorithmic processes, exacerbating societal inequalities and perpetuating systemic biases. The field of fairness in machine learning is a burgeoning area of research focused on preventing biases in data and model inaccuracies from resulting in unfavorable treatment of individuals based on characteristics such as race, gender, disabilities, and sexual or political orientation. It is important to address fairness and ethics in machine learning because the outcome can be detrimental to users and the community when machine learning isn’t fair. For example, algorithms on social media sites may have sparked political tensions due to skewed or siloed news feeds (and fake news), when the intention was to deliver personalized recommendations for users.
\vspace{.1cm}

To address the pressing need for fairness in machine learning, various techniques have been developed to mitigate unfairness for machine learning models at different stages of model development. These techniques are broadly categorized into pre-processing, in-processing, and post-processing methods \cite{wan2023processing}. Pre-processing techniques and post-processing methods offer straightforward strategies to address unfairness. Pre-processing methods focus on adjusting the training data distribution to balance sensitive groups, while post-processing methods calibrate prediction results post-model training. In contrast to the aforementioned techniques, in-processing debiasing methods have gained significant traction within the research community. These methods are noteworthy for their direct integration of fairness considerations into the model design process, resulting in the creation of intrinsically fair models \cite{dwork2012fairness, edwards2015censoring, hashimoto2018fairness, kearns2018preventing}. By doing so, they address fairness issues at a fundamental level within machine learning models, promoting fairness as an integral aspect of model architecture. This approach not only reflects the growing importance of fairness but also represents a substantial step towards the development of equitable and ethically sound machine learning systems. In-processing methods offer advantages such as directly considering fairness in model optimization, enabling the converged model to achieve fairness even with biased input data \cite{caton2020fairness}. Additionally, in-processing methods can effectively fine-tune representations from pre-trained models to mitigate bias without requiring extensive re-training efforts. Based on the stage at which fairness is achieved in the model, in-processing debiasing techniques can be categorized into explicit and implicit methods \cite{wan2023processing}. Explicit methods directly incorporate fairness constraints in training objectives, while implicit methods focus on refining latent representation learning. However, in our paper, we introduce a novel framework called "The Fairness Stitch" which combines the principles of model stitching with fairness constraint as an explicit in-processing debiasing method. This gives a comprehensive strategy to enhance fairness in deep learning models, providing a practical solution to mitigate biases and promote accurate outcomes at the same time.

The major contributions of the paper are as follows:
\begin{enumerate}
\item Proposing an innovative framework of "The Fairness Stitch" to better mitigate bias while preserving model performance.

\item Explaining the trade-off between fairness and performance within our proposed framework via information.
\item Empirically proving the limitation of last-layer fine-tuning to attain an optimal balance between fairness and performance.
\end{enumerate}
\vspace{.1cm}
The following sections are structured as follows: Section~\ref{Related_Work} discusses the literature review, Section~\ref{Preliminaries_Section} explains the mathematical notations and definitions used in the paper. Section~\ref{label_Fairness_Stitch} formally defines and elaborates on 'The Fairness Stitch' framework, offering an explanation for the design choice. Sections~\ref{Data Sources} and~\ref{Experiments} delineate the datasets and design of experiments. Finally, in Section~\ref{Results_Section}, we present the outcomes and findings achieved through the implementation of our framework.

\section{Related Work}\label{Related_Work}
De-biasing techniques are mainly categorized as (i) in-preprocessing (ii) pre-processing (iii) post-processing. In-processing de-biasing techniques in machine learning aimed at mitigating disparity were studied by M. Wan et al~\cite{wan2023processing}. They propose the use of adding regularizer to reduce the correlation between sensitive attributes and prediction outcomes. Contrary studies \cite{cherepanova2021technical} highlight that in-processing techniques are less effective for over-parameterized large neural networks, as these models can easily overfit fairness objectives during training, especially when the training data is imbalanced. This fairness overfitting issue raised by \cite{cherepanova2021technical} poses an open challenge. Over-parameterization of neural networks leads to highly flexible decision boundaries, and attempting to meet fairness criteria for one attribute can negatively impact fairness with respect to another sensitive attribute. However, over-parameterization has been essential for achieving high prediction accuracy, particularly in neural networks designed for challenging tasks. To address the problem of fairness criteria overfitting identified by \cite{cherepanova2021technical}, a novel framework proposed by \cite{mao2023last} called last-layer fairness fine-tuning. A similar line of research is conducted by T. Kamishima et al~\cite{kamishima2011fairness} by introducing prejudice index (PI) as a regularizer, which quantifies the level of dependence between a sensitive variable and a target variable. While R. Jiang et al~\cite{jiang2020wasserstein} used information geometry metric Wasserstein-1 distances between classifier outputs and sensitive information as a regularization in the optimization process. In a more practical setting, Alex Beutel et al \cite{beutel2019putting} propose a new metric of \textit{conditional equality } while implementing equality of opportunity. In the application of recommender systems Alex Beutel et al \cite{beutel2019fairness} propose pairwise comparisons from randomized experiments as a tractable way to measure fairness. \cite{zafar2019fairness, zafar2017fairness} comes with a novel measure of decision boundary (un)fairness. They used covariance between the sensitive attributes and the (signed) distance between the subjects’ feature vectors and the decision boundary as a metric for the classifier. Mao, Yuzhen, et al. \cite{mao2023last} have a different route and avoid adding fairness constraint as a regularization for deep neural networks model. This is because large deep neural models have the tendency to overfit fairness criteria. Instead, they used pre-training and fine-tune the last layer to fair trainning their model. 
Their proposed method involves training a model using empirical risk minimization and subsequently fine-tuning only the last layer with various fairness constraints as an in-processing de-biasing technique. Extensive experiments on benchmark image datasets with different fairness notions, the authors demonstrate the efficacy of last-layer fine-tuning in enhancing fairness. Other important work along this direction is~\cite{kirichenko2022last, lee2022surgical}. Here the authors have shown, that only fine-tuning the last layer(s) while keeping others layers frozen during the gradient descent achieves better performance. Last layer fine-tuning a pre-trained neural network on a smaller, more specific dataset achieves de-biasing. However in a recent counterclaim study by Kumar, Ananya, et al. \cite{kumar2022fine}, it was observed that fine-tuning actually can achieve worse accuracy than linear probing out-of-distribution (OOD). This is particularly in cases where the pretrained features are of high quality and the distribution shift is substantial. The authors conducted experiments across ten distinct distribution shift datasets, revealing that while fine-tuning tends to achieve, on average, a 2\% higher accuracy in-distribution (ID), it results in a 7\% lower accuracy when dealing with out-of-distribution (OOD) data, compared to the performance of linear probing. To address this disparity, they propose a two-step strategy known as LP-FT, which begins with linear probing and is subsequently followed by full fine-tuning. This approach capitalizes on the strengths of both fine-tuning and linear probing and consistently leads to superior performance. In their empirical evaluations, Kumar et al. \cite{kumar2022fine} demonstrated that LP-FT outperforms both fine-tuning and linear probing across various datasets, achieving 1\% higher accuracy in-distribution and an impressive 10\% higher accuracy out-of-distribution.
In this paper we propose  an innovative method \textit (model-stiching) for achieving fairness explained in \ref{label_Fairness_Stitch}. Our proposed method draws inspiration from the works of model stiching by Lenc et al.and  Bansal et al \cite{lenc2015understanding, bansal2021revisiting}. Model stitching is a tool to study the internal representations of deep neural models. Model stitching  combines the bottom layers of one pre-trained and frozen model (referred to as Model A) with the top layers of another model (referred to as Model B) using a trainable layer positioned in between, resulting in the creation of what is termed a "stitched model". Our work leveraged this model stitching to explore the commonalities and disparities in the learned representations of various models and training strategies as a strategy to mitigate bias.


\section{Preliminaries}\label{Preliminaries_Section}

\subsection{Notations}\label{Notation} We formalize our classification setting as a triplet dataset $T = \{ (x_{i}, a_{i}, y_{i}) \}_{i=1}^{N}$, $x_{i}$ is the feature drawn from an distribution over the alphabet $X$, $a_{i} \in A$ is sensitive attribute (race,
gender), and $y_{i} \in Y$ is the output label. The machine learning model is trained with standard loss functions of cross-entropy.

\begin{definition}[Group Fairness]
Group Fairness aims to ensure that different groups within a population are treated fairly, without discrimination, in the decision-making process. It involves assessing whether the outcomes or decisions produced by the algorithm are consistent and equitable across various predefined groups, typically distinguished by sensitive attributes (e.g., gender, race, age). Group fairness seeks to prevent bias or disparate impact on any particular group and promote fairness and equality in algorithmic decision-making. Mathematically, group fairness can be defined using various fairness metrics or statistical tests, depending on the specific context and objectives of a given study \cite{narayanan2018translation}. In our paper, our primary emphasis is on group fairness. We provide a succinct overview of three fairness definitions that are relevant to this paper. They serve as fairness constraints during the fine-tuning/training of our neural network model. We have adopted the same implementation as elucidated in~\cite{mao2023last} for fairness definitions. \\
\end{definition}
\begin{definition}[Equalized Odds]
Equalized Odds (EO) is a fairness criterion applied to classification tasks, ensuring equitable true positive and false positive rates between different groups. It's defined under the conditional probability distributions ($\mu_{0}$,  $\mu_{1}$), associated with distinct groups identified by sensitive attributes. A classifier $U$, satisfies equalized odds if, for all values of $y \in [0,1] $ , the following condition holds \cite{hardt2016equality}: $\mu_{0}(U(X) = 1 \mid Y = y) = \mu_{1}(U(X) = 1 \mid Y = y)$ for all $y \in [0,1]$. This ensures elimination disparities in different groups that are affected by both true positive (TPR) and false positive (FNR) rates in predictive models. To operationalize the pursuit of equalized odds, practitioners often employ a training strategy that involves minimizing specific objectives as ~\cite{cherepanova2021technical, padala2020fnnc}:
\begin{equation} \label{eq:equalized_odds}
	\min_{U} \bigl\{  \mathcal{L}_{w}(U) + \alpha (TPR + FNR) \bigl\}
\end{equation}
where $\alpha$ is the weight of classifier $U$, $\mathcal{L}$ signifies the cross-entropy loss, $p_i$ the softmax output.  and TPR and FNR are derived as:
\[ \Resize{6cm}{TPR = \left| \frac{\sum_{i} p_{i} (i - y_{i}) a_{i}}{\sum_{i} a_{i}} - \frac{\sum_{i} p_{i} (i - y_{i}) a_{i}}{\sum_{i} (1 - a_{i})} \right|}\]
\[ \Resize{6cm}{FNR = \left| \frac{\sum_{i} (i - p_{i}) y_{i} a_{i}}{\sum_{i} a_{i}} - \frac{\sum_{i} (1 - p_{i}) y_{i} (1 - a_{i})}{\sum_{i} (1 - a_{i})} \right| }\] 
\end{definition}
\begin{definition}[Accuracy equality]
Accuracy equality (AE)  evaluates whether the subjects in protected and unprotected groups experience similar False Positive Rate (FPR) and False Negative Rate (FNR). The objective of AE is to ensure that misclassification rates are approximately equal across these sensitive groups, such as different demographic categories, to prevent discriminatory or biased outcomes in predictive models~\cite{zafar2017fairness}.
\end{definition}

\begin{definition}[Max-Min Fairness]
Max-Min Fairness (MMF) is a fairness principle that prioritizes and maximizes the performance of the least advantaged or worse-off group within a given context ~\cite{rawls2001justice}. MMF  optimizes the performance for the group with the lowest utility while still meeting overall performance goals, used in the area of fairness-aware machine learning~\cite{lahoti2020fairness, cherepanova2021technical}. 
\end{definition}

\begin{definition}[Model Stitching]
Model Stitching (MS) is a method of combining the bottom-layers of a pre-trained ($A$) with the top-layers of another pre-trained model (B). This combination of top-layers with bottom-layers is done using a trainable layer in between and is termed as "Stitched Model" . For a neural network $A$, with architecture $\mathcal{A}$ and $r: \mathcal{X} \rightarrow \mathbb{R}^{d}$  be a representation, the loss function $\mathcal{L}$ is defined by~\cite{bansal2021revisiting}:
\begin{equation} \label{eq3}
	\begin{split}
		\mathcal{L}_{\ell}(r;A) = \inf_{s \in \mathcal{S}} \mathcal{L}(A_{>\ell}  \circ s \circ r)
	\end{split}
\end{equation}
where $\mathcal{S}$ is family of stitching layers, $A_{>\ell}$ mapping function from activations of $\ell$-th layer of $A$ to the final output, $\ell$ is the index of layer of $\mathcal{A}$ and  $\circ$ is function composition. $\mathcal{L}_{\ell}(r;A)$ is the minimum loss of stiching $r$ into all layers of $A$ except the initial $\ell$ layers, employing a stitching layer chosen from the set $\mathcal{S}$ as shown in Fig~\ref{fig:The-proposed-framework}. Model stitching tells us the path (homotopy) through the stitching layer $s$ of  transforming a candidate representation $r$ into
 first layers of $A$, in an appropiate subspace.

\begin{figure}
\centerline{\includegraphics[width=18.5pc]{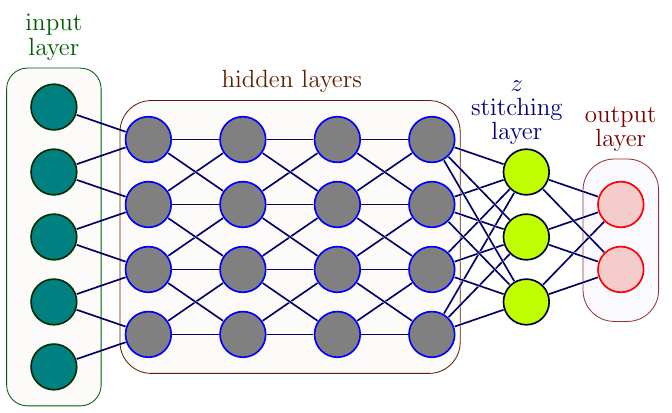}}
\caption{Illustration of 'The Fairness Stitch' framework for our stitched model. Our pre-trained model $\mathcal{M}$ includes only the input, hidden, and output layers, excluding the stitching layer $z$. The weights of the stitched model are kept frozen except for the weights associated with the stitching layer $z$.}
\label{fig:The-proposed-framework}
\end{figure}
\end{definition}


\subsection{Deep Neural Networks via Information}\label{Visualizing_DNNs}
In the context of deep neural networks, Shwartz-Ziv et al. \cite{shwartz2017opening} introduced the concept of the "information plane" to provide provable guarantees for network optimization through Stochastic Gradient Descent (SGD). This framework relies on the Mutual Information (MI) between random variables, denoted as $\Psi_{1}$ and $\Psi_{2}$, and it quantifies as $I (\Psi_{1}; \Psi_{2}) = H(\Psi_{1}) - H(\Psi_{1} \mid \Psi_{2})$. Here, $H(\Psi_{1})$ represents the entropy of $\Psi_{1}$, and $H(\Psi_{1} \mid \Psi_{2})$ denotes the conditional entropy of $\Psi_{1}$ given $\Psi_{2}$. Within the information plane, a representation variable denoted as $T$ is used to map input data $X$ and output labels $Y$, characterized by joint distributions $P(T{\mid}X)$ and $P(Y{\mid}T)$. Moreover, $T_{i}$, representing the $i^{th}$ hidden layer as a single multivariate variable, forming Directed Path Independence (DPI) chains \cite{shwartz2017opening}. 

As originally proposed by \cite{shwartz2017opening}, the examination of deep neural networks (DNNs) within the context of the "information plane" offers valuable insights. This visualization technique comes into play when the underlying distribution, denoted as $P(X;Y)$, is known, and when it's possible to calculate the encoder and decoder distributions, namely, $P(T{\mid}X)$, and $P (Y{\mid}T)$. Within this framework, two key order parameters, namely, $I(T;X)$ and $I(T;Y)$, serve as the means to visually compare different network architectures. Shwartz-Ziv et al. \cite{shwartz2017opening} also identified two key optimization phases through Stochastic Gradient Descent (SGD): the fast empirical error minimization (ERM) phase and the subsequent representation compression phase. During the ERM phase, spanning a few hundred epochs, layers notably increased their information on output labels $I_{Y} = I(T_{i};Y)$, $i \in 
 [1, ..., k]$, all while maintaining the Directed Path Independence (DPI) order, wherein lower layers retained higher label-related information. In the representation compression phase, the layers gradually reduced their information regarding input data $I_{X} = I(X;T_{i})$, shedding irrelevant information until convergence.

Shwartz-Ziv et al. \cite{shwartz2017opening} underscore that their analytical insights maintain broad applicability. They anticipate even more significant dynamic phase transitions for larger networks, a conjecture grounded in the statistical ensemble properties inherent to such networks. Taking inspiration from their confidence in the generalizability of their findings to "large real-world" problems and larger networks, we have reevaluated the concept of last-layer fairness fine-tuning, as discussed in Section \ref{label_rethinking}. Additionally, in Section \ref{Results_Section}, we provide empirical evidence that substantiates our reevaluation.

\section{Proposed Model}\label{label_Fairness_Stitch}
In this paper, we introduce "The Fairness Stitch" (TFS), a specialized layer aims to ensure equal opportunities for different groups during inference (based on sensitive attributes). "The Fairness Stitch" (TFS) transforms a candidate representation (a.k.a biased representation) into an unbiased representation. The unbiased representation comes from multiple sources, including unbiased pre-trained layers. Our "The Fairness Stitch" (TFS) acts as a path (homotopy) for transforming an unbiased representation to biased representation. Thus "The Fairness Stitch" (TFS) is used as a path to de-bias. By integrating "The Fairness Stitch," the machine learning model aims to achieve a more equitable representation of features, classes, and data points, ultimately enhancing overall fairness and minimizing disparities in the learning process. We test our method de-bising method by our proposed TFS, through comprehensive experiments on popular image datasets. We use different pre-trained architecture with different fairness notions, to demonstrate the efficacy of our framework in enhancing fairness. In our paper, we challenge the conventional notion of achieving fairness through last-layer fine-tuning \cite{mao2023last}. We instead show that freezing the last-layer is necessary and sufficient to strike a better balance between fairness and performance. Our proposed method "The Fairness Stitch" combines model stitching with fairness constraints as in-processing debiasing method (transformations from unfair representation to fair representation). Experiments validate that TFS achieves jointly fairness and accuracy.


\subsection{Rethinking Last-Layer Fairness Fine-Tuning}\label{label_rethinking}
Based on the generalization drawn from the findings discussed in the Section \ref{Visualizing_DNNs}, and when addressing neural network training with biased datasets, we can observe that during the representation compression phase, the final layer tends to contains significantly less information concerning the input data (which, in our case, is biased data). To elaborate further, as per the research conducted by \cite{shwartz2017opening}, during the representation compression phase, it is a common observation to find that $I(X;T_{j}) = \epsilon$ where $\epsilon$ is a very small positive number and $j$ refers to the last layers of the neural network (e.g. $j \in [n-3,n]$). This indicates that last layer of the model typically possesses minimal bias-related information from the input $X$. Consequently, using in-processing de-biasing methods such as applying fairness constraints for fine-tuning the pretrained last layer, should be approached cautiously and potentially reconsidered.

The last layer of a pretrained neural network is inherently designed for yielding high-performance predictions. However, any misguided attempt to intervene in the training process, such as solely fine-tuning the final layer, may result in underperforming on test examples sampled from out-of-distribution (OOD) data, especially when there is a significant distribution shift \cite{kumar2022fine}. Furthermore, this approach may disrupt the delicate balance between model performance and fairness considerations. Typically, in the representation compression phase, we aim for an optimal scenario where $0.8 \leq I(Y;T_{j}) \leq 1$ \cite{shwartz2017opening}. By focusing solely on debiasing the last layer, is akin to addressing the output of the last layer, resembling post-processing debiasing methods.

In fairness-aware in machine learning, it becomes essential to incorporate fairness constraints in updating the weights of the earlier layers throughout the two optimization phases since the value of $I(X;T_{i})$ is height in the representation compression phase, where $T_{i}$ indicates the earlier layers of the neural network. Regrettably, training the earlier layers with fairness constraints frequently yields unsatisfactory results \cite{cherepanova2021technical}. To surmount this challenge, we propose a novel approach called "The Fairness Stitch" that combines in-processing debiasing method with model stitching. Our approach involves introducing a trainable stitching layer between frozen layers in a deep learning model, thereby enhancing the trade-off between fairness and performance. For a comprehensive understanding of our framework, refer to Section \ref{Stitched_Last-Layers}. By avoiding fine-tuning the last layer, we hold the potential to enhance the trade-off between fairness and performance, a point underscored by the comparison between our findings in Section \ref{Experiments} and related work \cite{mao2023last}.

\subsection{Trainable Stitching Layer for Fairness}\label{Stitched_Last-Layers}
The deep learning architecture is formalized as a  composition of two distinct blocks: (i) the last layer (ii) preceding layers as shown in Figure ~\ref{fig:The-proposed-framework}. ''TFS'' is a two-step process. In the first phase, we train the model without the stitching layer denoted as $\mathcal{M}$. In the second phase, we add a trainable stitching layer between the two frozen layer blocks. In our paper we denote ${M}_{i}$ as the $i$-th layer of the pre-trained model. The cost of adding a stitching layer is given by:
\begin{equation} \label{eq4}
\begin{split}
\mathcal{L}^{*}(\mathcal{E}; z; r) = \inf_{z \in \mathcal{Z}} \left( \mathcal{L}(\mathcal{E} \circ z \circ r) + \textrm{fairness constraints} \right)
\end{split}
\end{equation}
where, $\mathcal{E} = \bigl\{ {M}_{i} \bigl\}_{i=0}^{n-1}$ is preceding layers, $r = \bigl\{{M}_{n} \bigl\}$ is last layer of  pre-trained model $\mathcal{M}$ and $z$ is stitching layer. The stitching layer $z$ is initialized with random weights and is trained by minimizing (\ref{eq4}). The added fairness constraints are explained in Section~\ref{Notation}. In our paper, the family of stitching layers $\mathcal{Z}$ sampled from the set of linear layers. The weights of the stitched model are frozen, except the stitching layer $z$ during the training. In summary, we train the stitching layer, that incorporates fairness constraints on the (class \& sensitive-attribute) balanced dataset.



\section{Data Sources and Characteristics:}\label{Data Sources}
The open-source CelebA and UTKFace datasets is used in the paper to design our experiments. We provide a summary of the dataset's characteristics, size, and the relevant attributes.\\

\textbf{CelebA Dataset}\footnote{\url{https://mmlab.ie.cuhk.edu.hk/projects/CelebA.html}}: It is a collection of celebrity faces, comprising more than 200,000 images annotated with 40 distinct attributes, including facial landmarks, gender, age, hair color, glasses, etc~\cite{liu2015deep}. For our experiments, we use hair color (blonde or non-blonde) as the target label ($y$), while gender (male or non-male) serves as the sensitive attribute ($a$). We follow the dataset division methodology outlined in~\cite{liu2015deep,mao2023last}, and create the Table \ref{table:CelebA_statistics}. To create a balanced sub-dataset, we perform sampling from the initial training and validation subsets based on the image count in the minority subgroup. Precisely, we select 1,569 images for each $(y, a)$ grouping, culminating in a total of 6,276 images within the balanced dataset. Furthermore (male, blonde hair) subgroup accounts for just $1\%$ of the total image count, signifying the minority category within the dataset.
\begin{table}[ht]
\centering
\begin{tabular}{|>{\raggedright\arraybackslash}p{0.55cm}|>{\centering\arraybackslash}p{2.0cm}|>{\centering\arraybackslash}p{2.3cm}|>{\raggedleft\arraybackslash}p{2.3cm}|}

\hline
  &  {\scriptsize Blonde Hair}  & {\scriptsize Non-blonde Hair}   &  {\scriptsize Total}\\ \hline

  {\scriptsize Male}  & {\scriptsize 1,387/182/180} & {\scriptsize 66,874/8,276/7,535} &  {\scriptsize 68,261/8,458/7,715}    \\ \hline

  {\scriptsize Female}  &  {\scriptsize 22,880/2,874/2,480} &{\scriptsize 71,629/8,535/9,767} & {\scriptsize 94,509/11,409/12,247}  \\ \hline

  {\scriptsize Total} &  {\scriptsize 24,267/3,056/2,660} &  {\scriptsize 138,503/16,811/17,302} & {\scriptsize 162,770/19,867/19,962}  \\ \hline
\end{tabular}
\caption{\textbf{(train/val/test)} Overview of the CelebA Dataset.}
\label{table:CelebA_statistics}
\end{table}

\textbf{UTKFace Dataset}\footnote{\url{https://susanqq.github.io/UTKFace/}}: It is a publicly accessible face dataset \cite{zhang2017age, mao2023last} that covers an extensive age range from newborns to individuals aged up to 116 years old. In our experiments, we use a random selection process to extract two subsets from the training dataset. Each subset accounts for 20\% of the original data, to maintain proportionality. One of these subsets served as the validation dataset, while the other as the test dataset as explained in Table~\ref{table:UTKFace_statistics}. In our experiment, age is used as the target variable, while gender is the sensitive attribute. Age is categorized into two groups: "young" ($ \leq 35$) and "others" ($ > 35$)~\cite{park2020readme}. We follow a similar procedure (as before) of constructing a balanced sub-dataset by sampling an equivalent number of images from both the original training and validation datasets for each distinct (y, a) group~\cite{mao2023last}. Each balanced group consists of  2,477 images, thereby yielding of 9,908 images in total and the minority category comprises females aged over 35 years.

\begin{table}[ht]
\centering
\begin{tabular}{|>{\raggedright\arraybackslash}p{0.55cm}|>{\centering\arraybackslash}p{2.0cm}|>{\centering\arraybackslash}p{2.3cm}|>{\raggedleft\arraybackslash}p{2.3cm}|}

\hline
 &  {\scriptsize Young } ($ \leq 35$)  &  {\scriptsize Old} ($ > 35$)   &  {\scriptsize Total}\\ \hline
		
{\scriptsize Male}  & {\scriptsize 4,133/1,378/1,378} & {\scriptsize 3,301/1,101/1,100} & {\scriptsize 7,434/2,479/2,478}   \\ \hline
		
{\scriptsize Female}  & {\scriptsize 4,931/1,643/1,644} & {\scriptsize 1,858/619/619} & {\scriptsize 6,789/2,262/2,263}   \\ \hline
		
{\scriptsize Total} & {\scriptsize 9,064/3,021/3,022} & {\scriptsize 5,159/1,720/1,719} & {\scriptsize 14,223/4,741/4,741}  \\ \hline
\end{tabular}
\caption{\textbf{(train/val/test)} Overview of the UTKFace Dataset.}
\label{table:UTKFace_statistics}
\end{table}

\section{Experimental Setup:}\label{Experiments}
This section explains the experimental setup to assess the efficacy of "The Fairness Stitch" (TFS), in promoting fairness. First, we introduce the baseline method (FDR). Subsequently, we provide detailed insights into the architectural specifications of the deep learning model used for both FDR and TFS. Lastly, we quantitatively evaluate the trade-off between performance and fairness with an array of performance metrics.

\subsection{Baseline Method}\label{Baselines_label}
This section explains (FDR) baseline method, for our experimental evaluations and acts as the point of reference for comparisons with our proposed framework (TFS). The FDR method introduced by Mao et al.~\cite{mao2023last}, uses empirical risk minimization (ERM), fine-tuning the last layer, and balanced dataset concerning both class and sensitive attributes. The balanced dataset is created through strategic sampling from the training and validation datasets. Furthermore, FDR augments its training process by incorporating fairness constraints into the objective function during fine-tuning. For a comprehensive understanding of the FDR method, we recommend referring to the original paper authored by Mao et al.~\cite{mao2023last}.

\subsection{Model Architecture}\label{models_label}
In this paper, we use ResNet-18 model~\cite{cherepanova2021technical} that is trained on CelebA and UTKFace datasets for facial recognition in both FDR and TFS. The ResNet-18 architecture consists of six blocks, including an input layer, and two convolutional layers within each of the four basic blocks. The final block corresponds to the last layer.

To ensure a harmonious alignment between the ResNet-18 model architecture and the conceptual framework of TFS illustrated in Figure~\ref{fig:The-proposed-framework}, we map the ResNet-18 model's input layer to the input layer within the conceptual TFS framework. Similarly, we correspond the four basic blocks in the ResNet-18 model to the hidden layers in the conceptual TFS framework. Furthermore, the final layer of the ResNet-18 model is matched with the last layer depicted in the conceptual TFS architecture presented in Figure~\ref{fig:The-proposed-framework}. In addition, we introduce a stitching layer denoted as $z$ between the hidden layers and the last layer of the ResNet-18 model, as visually demonstrated in Figure~\ref{fig:The-proposed-framework}. This stitching layer, $z$, is a fully connected layer designed with an input dimension matching that of the output from the hidden layers. Thus it produces an output dimension that matches the input dimension of the final layer. 

Both final models in TFS and FDR  frameworks utilize Stochastic Gradient Descent (SGD) with same specific hyperparameters, including a momentum of 0.9 and a weight decay of 5e-4.






\begin{figure}
	\centering
	\begin{subfigure}[b]{0.22\textwidth}
		\centering
		\includegraphics[width=\textwidth]{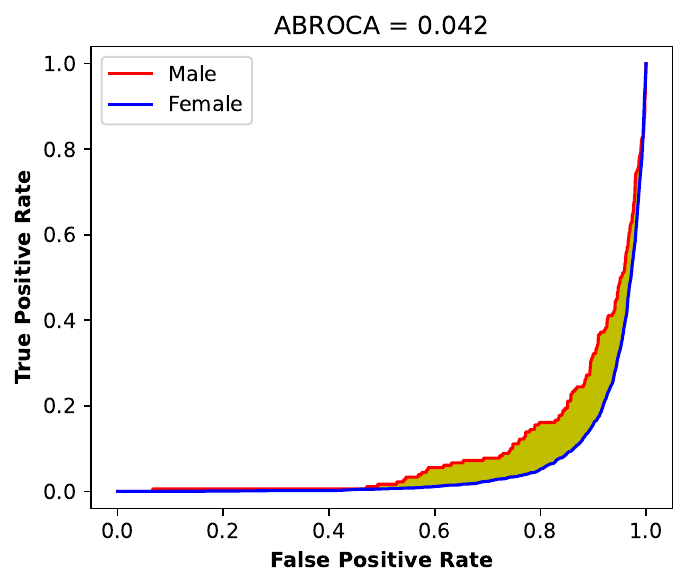}
		\caption{FDR \cite{mao2023last}}
		\label{fig:....554}
	\end{subfigure}
	\hfill
	\begin{subfigure}[b]{0.22\textwidth}
		\centering
		\includegraphics[width=\textwidth]{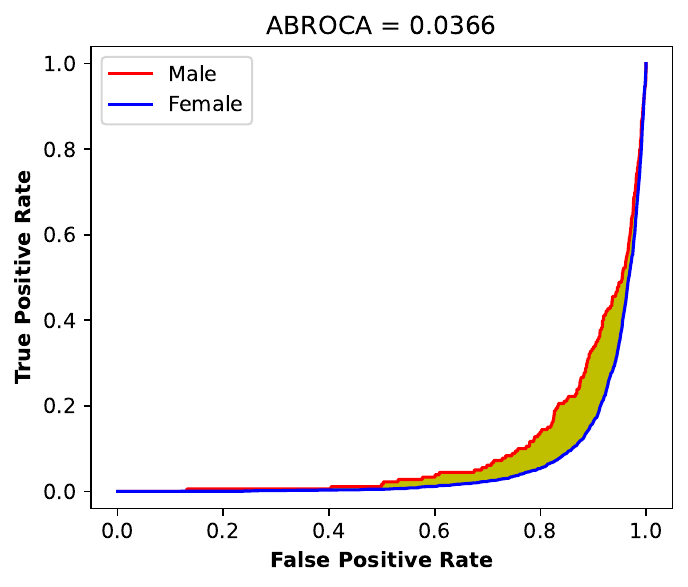}
		\caption{TFS}
		\label{fig:....22}
	\end{subfigure}
	\hfill
	\caption{ABROCA results on the CelebA. Figures (a) and (b) showcase ABROCA outcomes for both the FDR method \cite{mao2023last} and our TFS framework. Specifically, they showcase the use of equalized odds as the fairness constraint with $\alpha = 20$ during fine-tuning (for FDR) and training (for TFS), respectively.}
	\label{fig:ABROCA_CelebA}
\end{figure}

\begin{table*}[!ht]
\centering
\begin{tabular}{p{15.55cm}}
\hline
\textbf{Without Applying Any Fairness Constraint}  \\ \hline

\end{tabular}

\begin{tabular}{>{\centering\arraybackslash}p{3cm}|>{\centering\arraybackslash}p{3.28cm}|>{\centering\arraybackslash}p{3.28cm}|>{\centering\arraybackslash}p{1.28cm}|>{\centering\arraybackslash}p{1.28cm}|>{\centering\arraybackslash}p{1.28cm}} 
		
\hline
  &   BACC	 & AUC	 &  EO-Diff	 & AE & MMF \\ 
		
\end{tabular}

\begin{tabular}{>{\centering\arraybackslash}p{3cm}|>{\centering\arraybackslash}p{.7cm}|>{\centering\arraybackslash}p{1cm}|>{\centering\arraybackslash}p{.7cm}|>{\centering\arraybackslash}p{.7cm}|>{\centering\arraybackslash}p{1cm}|>{\centering\arraybackslash}p{.7cm}|>{\centering\arraybackslash}p{1.28cm}|>{\centering\arraybackslash}p{1.28cm}|>{\centering\arraybackslash}p{1.28cm}} \hline
			
 & {\scriptsize Train}	& {\scriptsize Balanced} & {\scriptsize Test}   &  {\scriptsize Train}	& {\scriptsize Balanced} & {\scriptsize Test}	& {\scriptsize Test}	& {\scriptsize Test} & {\scriptsize Test}    \\ \hline
		
Resnet18 Model  &     0.903	& 0.792 & 0.853  &   0.987	& 0.966 &  0.971 &   0.586 & 0.213 & 0.183	   \\
	
\hline
		
\end{tabular}
		
\begin{tabular}{p{15.55cm}}
\hline
\textbf{Fairness Notion 1: Equalized Odds} \\ \hline
			
\end{tabular}
	
\begin{tabular}{>{\centering\arraybackslash}p{3.0cm}|>{\centering\arraybackslash}p{3.28cm}|>{\centering\arraybackslash}p{3.28cm}|>{\centering\arraybackslash}p{3.28cm}|>{\centering\arraybackslash}p{1cm}} 
		
\hline
 &   BACC	 & AUC	 &  EO-Diff	 & AF  \\

 \end{tabular}
	
\begin{tabular}{>{\centering\arraybackslash}p{3cm}|>{\centering\arraybackslash}p{.7cm}|>{\centering\arraybackslash}p{1cm}|>{\centering\arraybackslash}p{.7cm}|>{\centering\arraybackslash}p{.7cm}|>{\centering\arraybackslash}p{1cm}|>{\centering\arraybackslash}p{.7cm}|>{\centering\arraybackslash}p{.7cm}|>{\centering\arraybackslash}p{1cm}|>{\centering\arraybackslash}p{.7cm}|>{\centering\arraybackslash}p{1cm}} \hline

	& {\scriptsize Train}	& {\scriptsize Balanced} & {\scriptsize Test}   &  {\scriptsize Train}	& {\scriptsize Balanced} & {\scriptsize Test}	& {\scriptsize Train}	& {\scriptsize Balanced} & {\scriptsize Test} &  {\scriptsize Test} 	   \\ \hline
		
	{\scriptsize FDR \cite{mao2023last}} &     0.898	& 0.896 & 0.876  &   0.959	& 0.956 & 0.942 &   0.014 & 0.015 & 0.110 &   0.766	   \\
		
		\hline
		
		TFS &     0.887	& 0.884 & 0.874   &   0.953	& 0.947 & 0.940 &  0.032	& 0.026 & 0.081 &   0.793	   \\
		
\hline
\end{tabular}

\begin{tabular}{p{15.55cm}}
		
\textbf{Fairness Notion 2: AE}   \\ \hline
		
\end{tabular}
	
\begin{tabular}{>{\centering\arraybackslash}p{3.0cm}|>{\centering\arraybackslash}p{3.28cm}|>{\centering\arraybackslash}p{3.28cm}|>{\centering\arraybackslash}p{3.28cm}|>{\centering\arraybackslash}p{1cm}} 
		 
		\hline
		&   BACC	 & AUC	 &  AE	 & AF  \\ 
\end{tabular}
	
\begin{tabular}{>{\centering\arraybackslash}p{3cm}|>{\centering\arraybackslash}p{.7cm}|>{\centering\arraybackslash}p{1cm}|>{\centering\arraybackslash}p{.7cm}|>{\centering\arraybackslash}p{.7cm}|>{\centering\arraybackslash}p{1cm}|>{\centering\arraybackslash}p{.7cm}|>{\centering\arraybackslash}p{.7cm}|>{\centering\arraybackslash}p{1cm}|>{\centering\arraybackslash}p{.7cm}|>{\centering\arraybackslash}p{1cm}} \hline
		
 & {\scriptsize Train}	& {\scriptsize Balanced} & {\scriptsize Test}   &  {\scriptsize Train}	& {\scriptsize Balanced} & {\scriptsize Test}	& {\scriptsize Train}	& {\scriptsize Balanced} & {\scriptsize Test} 	& {\scriptsize Test}  \\ \hline

{\scriptsize FDR \cite{mao2023last}}   &     0.906	& 0.905 & 0.883  &   0.967	& 0.964 & 0.949 &   0.009 & 0.001 & 0.008 &   0.875	   \\

\hline

 TFS   &     0.900	&  0.886 & 0.881  &   0.960	& 0.951 & 0.945 &   {\scriptsize 0.0002}	&  0.011 & {\scriptsize 0.0005} &   0.880	   \\ 
		
\hline
\end{tabular}

\begin{tabular}{p{15.55cm}}
		
		\textbf{Fairness Notion 3: MMF} \\ \hline
		
\end{tabular}

\begin{tabular}{>{\centering\arraybackslash}p{3.0cm}|>{\centering\arraybackslash}p{3.28cm}|>{\centering\arraybackslash}p{3.28cm}|>{\centering\arraybackslash}p{3.28cm}|>{\centering\arraybackslash}p{1cm}} 
		 
		\hline
		&   BACC	 & AUC	 &  WA	 & AF  \\

\end{tabular}

\begin{tabular}{>{\centering\arraybackslash}p{3cm}|>{\centering\arraybackslash}p{.7cm}|>{\centering\arraybackslash}p{1cm}|>{\centering\arraybackslash}p{.7cm}|>{\centering\arraybackslash}p{.7cm}|>{\centering\arraybackslash}p{1cm}|>{\centering\arraybackslash}p{.7cm}|>{\centering\arraybackslash}p{.7cm}|>{\centering\arraybackslash}p{1cm}|>{\centering\arraybackslash}p{.7cm}|>{\centering\arraybackslash}p{1cm}} \hline

		& {\scriptsize Train}	& {\scriptsize Balanced} & {\scriptsize Test}   &  {\scriptsize Train}	& {\scriptsize Balanced} & {\scriptsize Test}	& {\scriptsize Train}	& {\scriptsize Balanced} & {\scriptsize Test} 	& {\scriptsize Test} 	   \\ \hline

		{\scriptsize FDR \cite{mao2023last}}  &     0.915	& 0.913 & 0.875  &   0.978	& 0.972 & 0.96 & 0.864 	&  0.880 & 0.800 &   1.675	   \\ \hline
		
		\hline
		
		TFS  &     0.916	& 0.906 & 0.877 &   0.979	& 0.968 & 0.96 &  0.865	& 0.867 & 0.811 &   1.688		   \\ \hline

\end{tabular}

\caption{Results of our approach 'The Fairness Stitch' (TFS) with different fairness notions on CelebA dataset. For AUC, BACC, WA and AF, a larger value is considered better; while for EO-Diff and AE a smaller value is considered better.}
\label{table:TFS_Resnet18_Model_CelebA}
\end{table*}

\subsection{Performance and Fairness Metrics}\label{Metrics2}
In our experiments, we use a set of performance and fairness metrics to evaluate the effectiveness of our proposed framework (TFS). These metrics are instrumental in quantifying both the performance and fairness aspects of the models in both FDR and TFS.

\begin{itemize}[leftmargin=10pt]
\item \textbf{Balanced Accuracy (BACC)} \cite{brodersen2010balanced}: BACC measures the model's ability to correctly classify instances across different classes, considering the imbalance between classes. It provides a balanced view of the model's performance.
\item \textbf{Area under the ROC Curve (AUC)} \cite{fawcett2004roc}: AUC is used to evaluate a model's ability to discriminate between the positive and negative classes. A higher AUC indicates a better-performing model with stronger discrimination ability.

\item \textbf{Equalized Odds Difference} \cite{hardt2016equality} \textbf{(EO$\_$Diff)}: EO\_Diff quantifies disparities in the true positive rates (sensitivity or recall) and false positive rates (fallout or false alarm rate) between different demographic or protected groups. A smaller value of EO\_Diff indicates better fairness. 

\item \textbf{Accuracy Equality Difference} \textbf{(AE$\_$Diff)} \cite{mao2023last}: AE\_Diff measures the difference in misclassification rates between the two groups. A smaller value of AE\_Diff indicates better fairness, as it signifies that the model's predictive accuracy is more balanced between the groups, reducing the likelihood of one group being disadvantaged in terms of predictive accuracy compared to the other. 

\item  \textbf{Worst Accuracy} (WA) \cite{mao2023last}: WA assesses the minimum accuracy among various group combinations, with a larger value indicating better fairness. 

\item \textbf{AF}: The AF metric, introduced by Mao et al. in their work \cite{mao2023last}, combines both balanced accuracy (BACC) and a fairness metric to holistic assess a machine learning model. A larger AF value indicates better fairness.  It's calculated as AF = BACC - EO\_Diff (for Equalized Odds), AF = BACC - AE\_Diff (for Accuracy Equality), and AF = BACC + WA (for Max-Min Fairness).



\item \textbf{Absolute Between-ROC Area} (ABROCA) \cite{gardner2019evaluating}: ABROCA is based on the Receiver Operating Characteristics (ROC) and quantifies the discrepancy between the baseline and comparison group curves across all potential thresholds. It accomplishes this by summing up this discrepancy while disregarding which subgroup's model performs better at specific thresholds, as it considers the absolute values of these differences. ABROCA typically falls within the range of 0 to 1, but in practical scenarios, it often resides within the interval [0.5, 1]. 
\end{itemize}

\section{Results:}\label{Results_Section}
This section explains and provides the comparative evaluation results of 'TFS' and 'FDR'. We fine-tune our models for 1000 epochs using (class \& sensitive-attribute) balanced dataset for the evaluation. Our objective is to assess the potential of both FDR and TFS in achieving an optimal trade-off between fairness and performance. We first select the best fine-tuned model, both from FDR and TFS, based on their performance during the 1000 epochs of fine-tuning on the validation dataset. Subsequently, we present our findings by testing the best fine-tuned TFS and FDR models on the test dataset. Our results affirm that TFS outperforms FDR in achieving a superior balance between fairness and performance.

\subsection{Results without Applying De-biasing Technique}
Our research findings, as shown in Table~\ref{table:TFS_Resnet18_Model_CelebA} and Table~\ref{table:TFS_Resnet18_Model_CelebA} \ref{table:TFS_Resnet18_Model_UTKFace}, reveals a distinct disparity in bias levels between the CelebA and UTKFace datasets before using of any debiasing techniques. Specifically, the CelebA dataset exhibits significantly higher bias ratios, with EO-Diff = 0.586, AE = 0.213, and MMF = 0.183. In contrast, the UTKFace dataset demonstrates lower bias ratios, with EO-Diff = 0.143, AE = 0.023, and MMF = 0.675. This difference can be attributed to the varying distribution of samples among different sensitive attributes, particularly in terms of gender. The CelebA dataset contains a higher disparity in the number of male and female samples. Additionally, it's essential to emphasize that the CelebA dataset is considerably larger in size compared to the UTKFace dataset, emphasizing the need for dataset-specific adjustments in hyperparameters. For instance, in our subsequent experiments with FDR and TFS on the CelebA dataset, we set $\alpha$ to 20 for the EO and AE constraints. Contrary, in fine-tuning with the UTKFace dataset, we opted for a value of $\alpha=2$ for the EO and AE constraints. The differences highlight the importance of tailored hyper-parameter selection based on dataset characteristics.

\subsection{Results with Applying TFS }\label{TFS_Resnet18_Model_CelebA_UTKFace}
\begin{figure}
	\centering
	\hfill
	\begin{subfigure}[b]{0.22\textwidth}
		\centering
		\includegraphics[width=\textwidth]{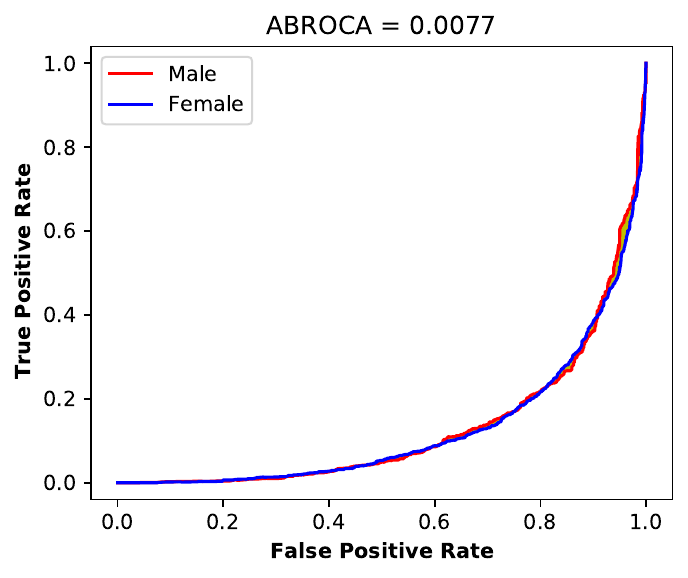}
		\caption{FDR \cite{mao2023last}}
		\label{fig:.444.}
	\end{subfigure}
	\hfill
	\begin{subfigure}[b]{0.22\textwidth}
		\centering
		\includegraphics[width=\textwidth]{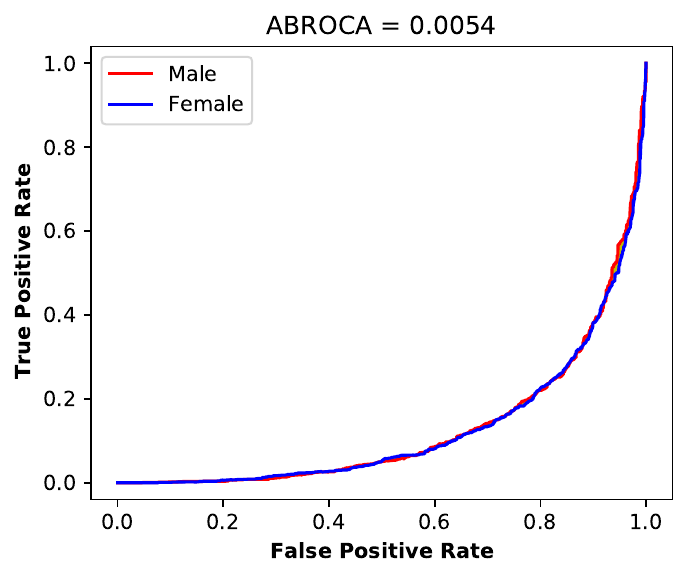}
		\caption{TFS}
		\label{fig:33.}
	\end{subfigure}
	\caption{ABROCA results on the UTKFace. Figures (a) and (b) showcase ABROCA outcomes for both the FDR method \cite{mao2023last} and our TFS framework. Specifically, they showcase the use of equalized odds as the fairness constraint with $\alpha = 2$ during fine-tuning (for FDR) and training (for TFS), respectively.}
	\label{fig:ABROCA_UTKFace}
\end{figure}

In this section we present the results obtained by applying TFS framework to a Resnet18 model for binary classification task within the CelebA and UTKFace datasets. Our research findings shed light on the performance and effectiveness of TFS in promoting fairness while maintaining accurate predictions. We summarize the outcomes of applying the three distinct fairness criteria within the TFS framework to the CelebA and UTKFace datasets in Tables \ref{table:TFS_Resnet18_Model_CelebA} and \ref{table:TFS_Resnet18_Model_UTKFace} as follows:

Our experimental findings highlight the superiority of TFS framework over FDR method when evaluating their performance and fairness under the equalized odds fairness constraint. TFS consistently achieves significantly lower EO-Diff fairness values for both the CelebA and UTKFace datasets. Specifically, on the CelebA test dataset, FDR records an EO-Diff value of 0.110, whereas the TFS framework notably reduces this to 0.081. Similarly, on the UTKFace test dataset, FDR exhibits an EO-Diff value of 0.062, while TFS further diminishes it to 0.058. Furthermore, Figures \ref{fig:ABROCA_CelebA} and \ref{fig:ABROCA_UTKFace} illustrate the ABROCA values, providing a comprehensive comparison of the performance and fairness achieved by the FDR and TFS frameworks when equalized odds is employed as a fairness constraint. These figures unequivocally demonstrate the superior fairness of our TFS framework (ABROCA = 0.0366) over FDR (ABROCA = 0.042). Furthermore, TFS exhibits minimal and negligible deviations on BACC and AUC metrics when equalized odds. This is used as a fairness constraint during training on both the CelebA and UTKFace datasets. On the CelebA test dataset, our TFS framework achieves a BACC of 0.874 and an AUC of 0.940, while the FDR method records values of 0.876 for BACC and 0.942 for AUC. These slight variations underscore the consistency and reliability of the TFS framework in delivering both fairness and accuracy. In addition, when we consider accuracy equality as a fairness constraint for both the CelebA and UTKFace datasets, TFS consistently achieves lower AE value compared to FDR. On the CelebA test dataset, TFS reduces the AE value from 0.008 (FDR) to an impressive 0.0005, emphasizing its robust capability to promote accuracy equality. Similarly, on the UTKFace dataset, FDR records an AE value of 0.016, whereas TFS further diminishes it to 0.0096. Moreover,  TFS shows stability with minimal and inconsequential deviations in terms of both BACC and AUC metrics when accuracy equality is employed as a fairness constraint on both the CelebA and UTKFace datasets. Enforcing accuracy equality as a fairness constraint for CelebA dataset, FDR framework consistently achieves a BACC of 0.883 and an AUC of 0.949, while the TFS method yields values of 0.881 for BACC and 0.945 for AUC. These negligible variations emphasize the reliability and consistency of the TFS framework in simultaneously optimizing fairness and accuracy. Finally, under the MMF constraint, TFS achieves higher WA values, indicating improved performance for both datasets. On the CelebA dataset, TFS achieves a WA value of 0.811, surpassing FDR's 0.800. Similarly, on the UTKFace dataset, TFS attains a WA value of 0.744, outperforming FDR's 0.739. Moreover, when employing MMF as a fairness constraint, TFS exhibits minimal and negligible deviations in BACC and AUC metrics on both datasets, reinforcing its ability to strike a favorable balance between fairness and performance. Overall, these results underscore the effectiveness of our TFS framework in achieving an optimal balance between performance and fairness, as demonstrated in Table ~\ref{table:TFS_Resnet18_Model_CelebA} and Table~\ref{table:TFS_Resnet18_Model_UTKFace}, compared to the baseline method.

\begin{table*}[ht]
	\centering

	\begin{tabular}{p{15.55cm}}
		\hline
		\textbf{Without Applying Any Fairness Constraint}  \\ \hline

	\end{tabular}

	\begin{tabular}{>{\centering\arraybackslash}p{3cm}|>{\centering\arraybackslash}p{3.28cm}|>{\centering\arraybackslash}p{3.28cm}|>{\centering\arraybackslash}p{1.28cm}|>{\centering\arraybackslash}p{1.28cm}|>{\centering\arraybackslash}p{1.28cm}} 
		
		\hline
		&   BACC	 & AUC	 &  EO-Diff	 & AE & MMF \\

	\end{tabular}

	\begin{tabular}{>{\centering\arraybackslash}p{3cm}|>{\centering\arraybackslash}p{.7cm}|>{\centering\arraybackslash}p{1cm}|>{\centering\arraybackslash}p{.7cm}|>{\centering\arraybackslash}p{.7cm}|>{\centering\arraybackslash}p{1cm}|>{\centering\arraybackslash}p{.7cm}|>{\centering\arraybackslash}p{1.28cm}|>{\centering\arraybackslash}p{1.28cm}|>{\centering\arraybackslash}p{1.28cm}} \hline

		& {\scriptsize Train}	& {\scriptsize Balanced} & {\scriptsize Test}   &  {\scriptsize Train}	& {\scriptsize Balanced} & {\scriptsize Test}	& {\scriptsize Test}	& {\scriptsize Test} & {\scriptsize Test}    \\ \hline
		
		Resnet18 Model  &     0.998	& 0.947 & 0.803  &   1.000	& 0.983 &  0.885 &   0.143 & 0.023 & 0.675	   \\

		\hline
		
	\end{tabular}

	\begin{tabular}{p{15.55cm}}
		\hline
		\textbf{Fairness Notion 1: Equalized Odds}   \\ \hline
	\end{tabular}
	
	\begin{tabular}{>{\centering\arraybackslash}p{3.0cm}|>{\centering\arraybackslash}p{3.28cm}|>{\centering\arraybackslash}p{3.28cm}|>{\centering\arraybackslash}p{3.28cm}|>{\centering\arraybackslash}p{1.03cm}} 
		
		\hline
		&   BACC	 & AUC	 &  EO-Diff	 & AF  \\ 
		
	\end{tabular}
	
	\begin{tabular}{>{\centering\arraybackslash}p{3cm}|>{\centering\arraybackslash}p{.7cm}|>{\centering\arraybackslash}p{1cm}|>{\centering\arraybackslash}p{.7cm}|>{\centering\arraybackslash}p{.7cm}|>{\centering\arraybackslash}p{1cm}|>{\centering\arraybackslash}p{.7cm}|>{\centering\arraybackslash}p{.7cm}|>{\centering\arraybackslash}p{1cm}|>{\centering\arraybackslash}p{.7cm}|>{\centering\arraybackslash}p{1cm}} \hline
		
		& {\scriptsize Train}	& {\scriptsize Balanced} & {\scriptsize Test}   &  {\scriptsize Train}	& {\scriptsize Balanced} & {\scriptsize Test}	& {\scriptsize Train}	& {\scriptsize Balanced} & {\scriptsize Test} &  {\scriptsize Test} 	   \\ \hline
		
		{\scriptsize FDR \cite{mao2023last}} &     0.996	& 0.951 & 0.796  &   1.000	& 0.986 & 0.876 &   0.005 & 0.010 & 0.062 &   0.734	   \\
		
		\hline
		
		TFS &     0.996	& 0.951 & 0.793  &   1.000 & 0.985 & 0.876 &  0.003 & 0.012 & 0.058 &   0.735	   \\
		\hline
  
		\hline
	\end{tabular}
	
	\begin{tabular}{p{15.55cm}}
		
		\textbf{Fairness Notion 2: AE}   \\ \hline
		
	\end{tabular}
	
	\begin{tabular}{>{\centering\arraybackslash}p{3.0cm}|>{\centering\arraybackslash}p{3.28cm}|>{\centering\arraybackslash}p{3.28cm}|>{\centering\arraybackslash}p{3.28cm}|>{\centering\arraybackslash}p{1.03cm}} 
		
		\hline
		&   BACC	 & AUC	 &  AE	 & AF  \\ 
	\end{tabular}
	
	\begin{tabular}{>{\centering\arraybackslash}p{3cm}|>{\centering\arraybackslash}p{.7cm}|>{\centering\arraybackslash}p{1cm}|>{\centering\arraybackslash}p{.7cm}|>{\centering\arraybackslash}p{.7cm}|>{\centering\arraybackslash}p{1cm}|>{\centering\arraybackslash}p{.7cm}|>{\centering\arraybackslash}p{.7cm}|>{\centering\arraybackslash}p{1cm}|>{\centering\arraybackslash}p{.7cm}|>{\centering\arraybackslash}p{1cm}} \hline

		& {\scriptsize Train}	& {\scriptsize Balanced} & {\scriptsize Test}   &  {\scriptsize Train}	& {\scriptsize Balanced} & {\scriptsize Test}	& {\scriptsize Train}	& {\scriptsize Balanced} & {\scriptsize Test} 	& {\scriptsize Test}  \\ \hline

		{\scriptsize FDR \cite{mao2023last}}    &  0.998	& 0.948 & 0.798 &  1.000	& 0.985 & 0.884 &  0.0008 & 0.003 & 0.016 &   0.782	 	   \\

		\hline

		TFS   &  0.992 & 0.946 & 0.796 & 1.000 & 0.984 &  0.879 &  0.010 & 0.001 & 0.0096 & 0.7864 \\

		\hline
		
		\hline

	\end{tabular}

	\begin{tabular}{p{15.55cm}}
		
		\textbf{Fairness Notion 3: MMF}  \\ 
		
		\hline
		
	\end{tabular}

	\begin{tabular}{>{\centering\arraybackslash}p{3cm}|>{\centering\arraybackslash}p{3.28cm}|>{\centering\arraybackslash}p{3.28cm}|>{\centering\arraybackslash}p{3.28cm}|>{\centering\arraybackslash}p{1cm}} 
		
		\hline
		&   BACC	 & AUC	 &  WA	 & AF  \\

	\end{tabular}

	\begin{tabular}{>{\centering\arraybackslash}p{3cm}|>{\centering\arraybackslash}p{.7cm}|>{\centering\arraybackslash}p{1cm}|>{\centering\arraybackslash}p{.7cm}|>{\centering\arraybackslash}p{.7cm}|>{\centering\arraybackslash}p{1cm}|>{\centering\arraybackslash}p{.7cm}|>{\centering\arraybackslash}p{.7cm}|>{\centering\arraybackslash}p{1cm}|>{\centering\arraybackslash}p{.7cm}|>{\centering\arraybackslash}p{1cm}} \hline
		
		
		& {\scriptsize Train}	& {\scriptsize Balanced} & {\scriptsize Test}   &  {\scriptsize Train}	& {\scriptsize Balanced} & {\scriptsize Test}	& {\scriptsize Train}	& {\scriptsize Balanced} & {\scriptsize Test} 	& {\scriptsize Test} 	   \\ \hline

		{\scriptsize FDR \cite{mao2023last}}  & 0.998 & 0.949 & 0.797 &    1.000	& 0.984 & 0.881 &  0.997	& 0.933 & 0.739 &   1.536		   \\ \hline
		
		\hline
		
		TFS  &     0.997	& 0.949 & 0.797 &   1.000	& 0.984 & 0.880 &  0.995	& 0.934 & 0.744 &   1.541		   \\ \hline

	\end{tabular}

	\caption{Results of our approach 'The Fairness Stitch' (TFS) with different fairness notions on UTKFace dataset. For AUC, BACC, WA and AF, a larger value is considered better; while for EO-Diff and AE a smaller value is considered better.}
	\label{table:TFS_Resnet18_Model_UTKFace}
\end{table*}

\begin{figure}
	\centering
	\begin{subfigure}[b]{0.23\textwidth}
		\centering
		\includegraphics[width=\textwidth]{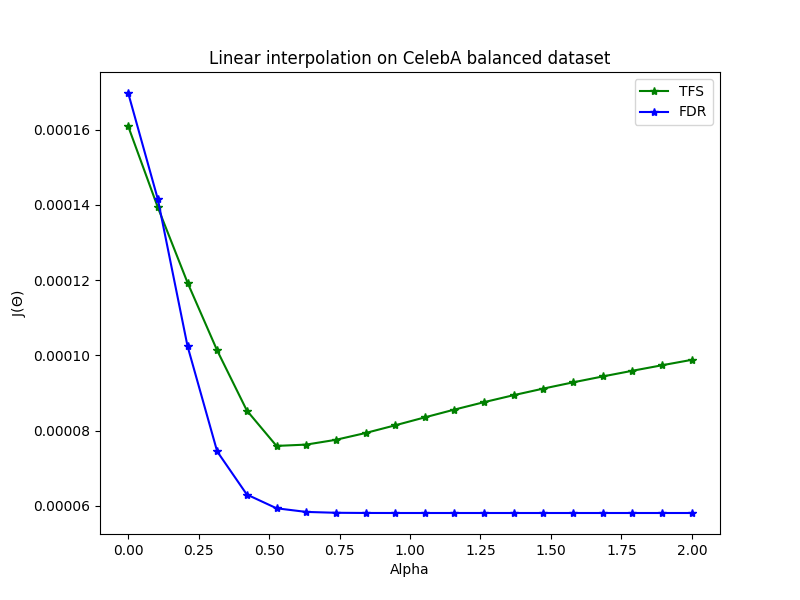}
		\caption{Loss on CelebA Balanced Dataset}
		\label{fig:CelebA_Balanced_dataset_loss}
	\end{subfigure}
	\hfill
	\begin{subfigure}[b]{0.23\textwidth}
		\centering
		\includegraphics[width=\textwidth]{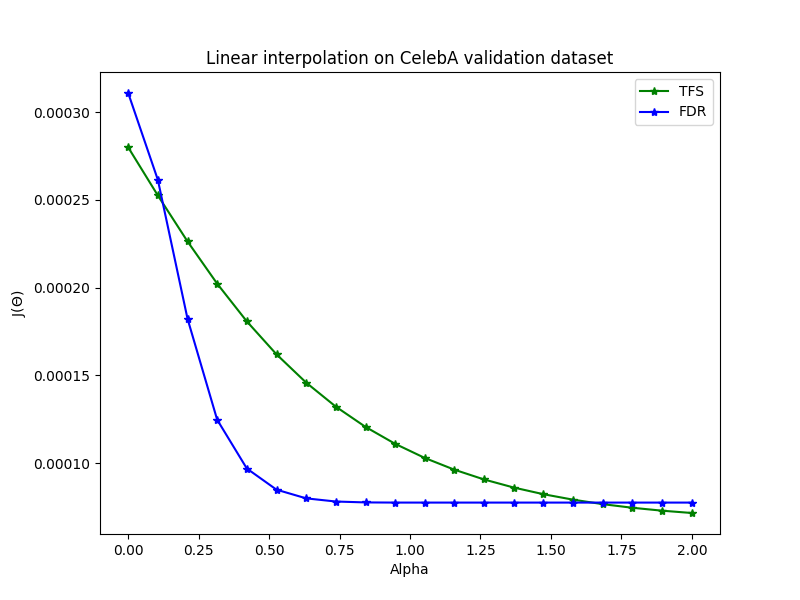}
		\caption{Loss on CelebA Validation Dataset}
		\label{fig:CelebA_Validation_dataset_loss}
	\end{subfigure}
	\hfill
	\caption{The linear interpolation curves for Resnet18 model using TFS and FDR frameworks on CelebA balanced and validation datasets.}
	\label{fig:loss2_CelebA}
\end{figure}

\subsection{Loss Function Visualization}\label{1DInterpolation}
To discern the distinctions between our proposed framework, TFS, and the baseline method FDR, we employ the 1-dimensional linear interpolation approach outlined in Goodfellow et al.'s work~\cite{goodfellow2014qualitatively}. This technique allows us to visualize and compare the objective functions of both methods, facilitating a comprehensive analysis of their performance. To visualize the loss function using 1-dimensional linear interpolation approach, we select two parameter vectors, denoted as $\theta_{0}$ and $\theta^{*}$, and chart the loss function values along the line that connects these two points. This line can be parametrized by selecting a scalar parameter, denoted as Alpha ($\alpha$), and establishing the weighted average as $\theta(\alpha) = (1-\alpha) \theta_{0} + \alpha \theta^{*}$. Here, $\theta^{*}$ represents the parameters (weights) of the trained/fine-tuned model, whether employing the TFS or FDR method, while $\theta_{0}$ signifies the parameters (weights) of the corresponding model in its randomly initialized state, before undergoing the training or fine-tuning process. In Figure \ref{fig:loss2_CelebA}, we present a visual representation of the function $J(\theta)$, defined as $J(\theta) = \mathcal{L}(\theta(\alpha))$, evaluated on both the balanced and validation CelebA datasets, where $\mathcal{L}$ represents the loss function employed within the TFS and FDR frameworks. Figure \ref{fig:loss2_CelebA} serves to underscore the differentiation between the TFS and FDR methods. Notably, our results indicate that FDR consistently outperforms TFS in terms of $J(\theta)$ on the balanced CelebA datasets. However, on the validation CelebA dataset, TFS consistently exhibits a better value at the final stage, signifying a more favorable trade-off between fairness and performance when compared to the FDR method. Furthermore, the outcomes depicted in Figure \ref{fig:loss2_CelebA} align with the findings presented in Table \ref{table:TFS_Resnet18_Model_CelebA}. Our observations show that while FDR surpasses TFS on the balanced dataset, the situation is different when considering the validation and test datasets. This suggests that our TFS framework has the capacity to generalize more effectively, striking a superior balance between fairness and performance compared to the FDR baseline method.


\section{Conclusion and Future Work }
In this paper, we have presented an innovative method "The Fairness Stitch (TFS)" as a de-biasing technique. TFS combines model stitching and fairness constraints to mitigate the risk of bias. We test the efficacy of our method by testing on two popular open-source datasets CelebA and UTKFace. We compare our results with the baseline method. Our research findings show TFS beats the baseline method in fairness and accuracy. Our work shows a practical de-biasing method with respect to computational complexity and sample complexity, especially in the deep learning models. We hope our method will help to contribute in advancing the active research area of fairness-aware machine learning. Our proposed method poses a challenge to the conventional wisdom of the effectiveness of the last layer in mitigating bias. 'TFS' complements the surgical fine-tuning \cite{lee2022surgical} in the fairness context and provokes us to rethink of the efficacy of the last layer.


In our future research work, we aim to investigate the computational and sample complexity of our TFS method with different non-linear stitchable layers on different pre-trained substrates. We will study the trade-off between fairness metrics with computational budget. The other research direction we plan to extend our TFS in generative adversarial networks.




\section*{Acknowledgment}
This research has been funded by European Social Fund via IT Academy programme. Additionally, this research is supported by The High Performance Computing Center at the University of Tartu, Estonia.

\bibliographystyle{IEEEtran}
\bibliography{TAI_template}

\begin{IEEEbiography}[{\includegraphics[width=1in,height=1.25in,clip,keepaspectratio]{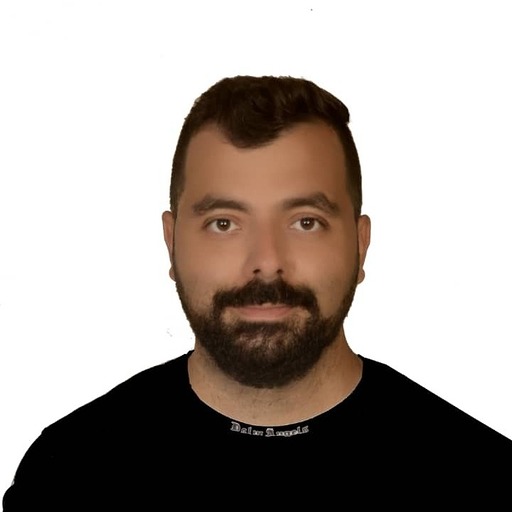}}] {Modar Sulaiman} is currently a Junior Research Fellow in Artificial Intelligence at the Institute of Computer Science, University of Tartu, Estonia, where he is also pursuing his Ph.D. in Computer Science. He holds a double master’s degree from InterMaths.EU program in 2017: MSc in Mathematical Engineering with a specialization in Scientific Computing from the Università dell’Aquila, Italy, and another MSc in Mathematics with a focus on Mathematical Modelling from Uniwersytet Śląski w Katowicach, Poland. In 2013, he earned his Bachelor of Science degree in Mathematics from Tishreen University, Syria. His research interests include fairness in machine learning, AI Ethics, and applied mathematics.

\end{IEEEbiography}

\begin{IEEEbiography}[{\includegraphics[width=1in,height=1.25in,clip,keepaspectratio]{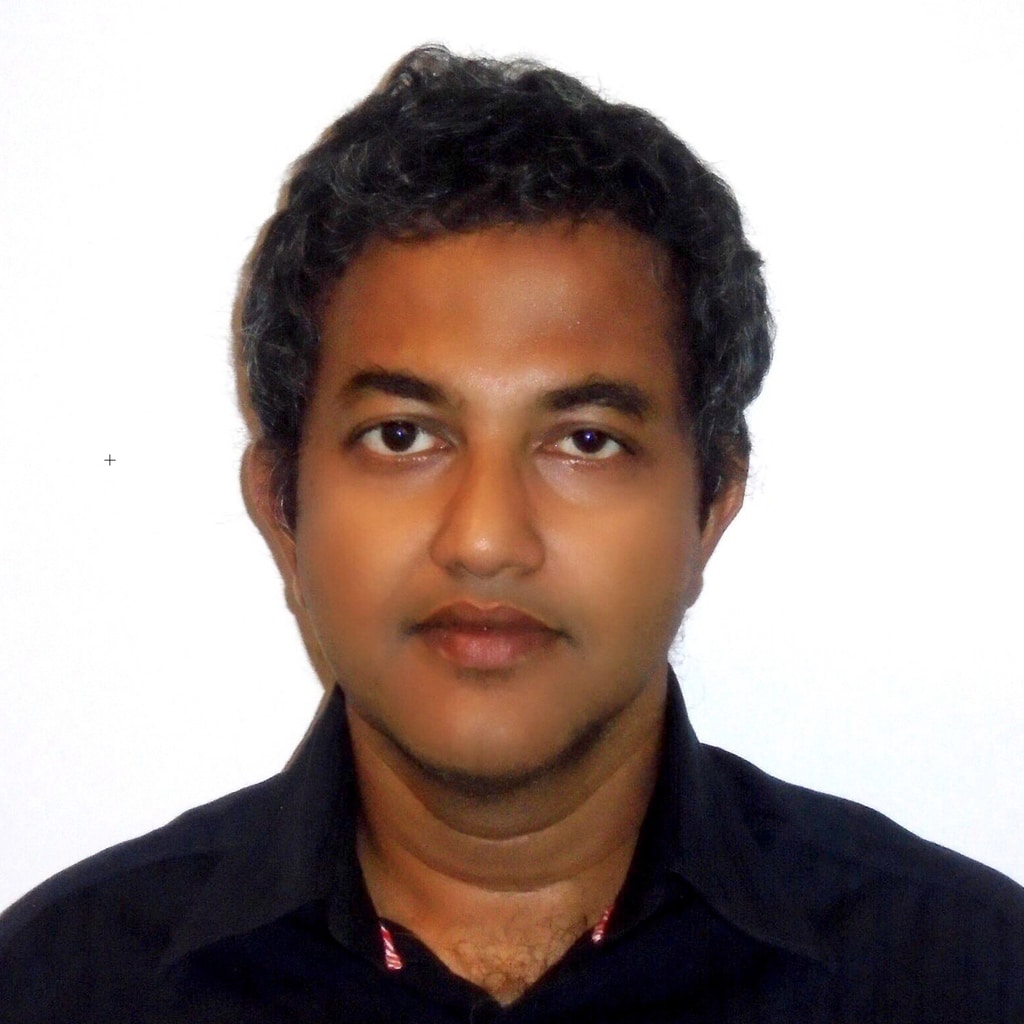}}]{Kallol Roy} is currently working as an Assistant Professor at Institute of ComputerScience, University of Tartu, Estonia. His research interest is deep Learning, computer vision, AI accelerator, and electromagnetics. He worked as a postdoctoral researcher at Packaging Research Center, Georgia Institute of Technology, Atlanta, USA and Statistical Artificial Intelligence Lab, Ulsan National Institute of Technology, South Korea and at Department of Mathematics, Indian Institute of Science Bangalore. He did his Bachelors in Electrical Engineering from Indian Institute of Technology (IIT K), Kanpur and PhD in Electrical Communication Engineering from Indian Institute of Science (IISc) Bangalore. He is a recipient of APS-IUSSTF Physics Student Visitation Award, 2012 Microsoft Travel Award, Sterlite Best Paper Award at Photonics 2010, IIT Guwahati, MHRD Scholarship, Government of India 2007, Jawaharlal Nehru Scholarship Steel Authority of India Limited, 2000.

\end{IEEEbiography}

\end{document}